\documentclass{article}

\usepackage{spconf,amsmath,amssymb,graphicx,hyperref}
\usepackage{algorithm}
\usepackage{algorithmic}
\usepackage{tabularx}
\usepackage{array}

\title{RSKD: Grounding Semantic Consistency in Driving VLMs via Distillation}
\name{
Chuchu Xie$^{1}$,
Yang Zhong$^{2}$,
Kangan Qian$^{1}$,
Hao Ye$^{2}$,
Kun Jiang$^{1}$,
Diange Yang$^{1}$
}

\address{
$^{1}$ School of Vehicle and Mobility, Tsinghua University, Beijing, China \\
$^{2}$ Automotive and Robotics, Xiaomi Corporation, Beijing, China \\
$^{1}$ \texttt{xcc22@mails.tsinghua.edu.cn},
\texttt{qka8203191007@163.com} \\
$^{1}$ \texttt{jiangkun@tsinghua.edu.cn},
\texttt{ydg@tsinghua.edu.cn} \\
$^{2}$ \texttt{zhongyang3@xiaomi.com},
\texttt{rawk.vx@gmail.com}
}
\begin{document}
%
\maketitle
\begin{abstract}
Knowledge distillation can transfer large multimodal models to compact models for autonomous driving, but uniform token-level matching can dilute planning-relevant semantics and amplify teacher errors. We propose RSKD, a semantic-consistency-guided distillation framework that routes supervision according to both the task importance of an answer phrase and its semantic consistency with the teacher prediction. Task-critical semantic phrases are localized in the ground-truth answer, and a phrase-level consistency score controls the balance between Jensen--Shannon divergence (JSD) distillation and ground-truth supervision. Teacher information is emphasized only where it is semantically consistent, while semantically inconsistent phrases receive stronger ground-truth supervision. We evaluate RSKD on DrivingVQA, LingoQA, and MapLMv2. RSKD improves over the compared student-based baselines on DrivingVQA and LingoQA and remains competitive on MapLMv2. 
\end{abstract}
\section{Introduction}
\label{sec:intro}
Multimodal large language models (MLLMs) have recently shown promising capabilities in autonomous driving, supporting scene understanding, visual question answering, and reasoning over complex traffic environments \cite{sima2023drivelm,tian2024drivevlm,wang2024omnidrive}.
Despite their strong reasoning capabilities, deploying large-scale MLLMs in practical autonomous systems remains challenging due to their substantial computational and memory costs. Knowledge distillation (KD) provides an effective approach to transferring knowledge from a powerful teacher model to a compact student model
\cite{hinton2015distilling,gu2024minillm,agarwal2024gkd}.

However, effective knowledge transfer in autonomous driving requires more
than uniformly matching teacher distributions across answer tokens.
Driving-oriented answers contain heterogeneous semantic information:
some tokens encode task-critical elements such as traffic participants,
traffic states, spatial relations, road constraints, and driving behaviors,
whereas others mainly serve descriptive or linguistic purposes.
Consequently, treating all answer positions identically may dilute
supervision on planning-relevant knowledge.

Recent studies have therefore explored fine-grained knowledge distillation
by adapting supervision according to token-level learning signals
\cite{jung2025todi,xie2026adakd}.
While such strategies improve the granularity of knowledge transfer,
model-derived token statistics do not necessarily indicate whether the
corresponding tokens encode task-critical driving semantics.
Semantic importance alone, however, does not determine how strongly the teacher should supervise a task-critical semantic phrase, which the teacher may still predict incorrectly.

Teacher confidence does not necessarily imply semantic correctness; a teacher
may confidently predict an incorrect spatial relation, traffic state, or
driving constraint. Strengthening distribution matching in such regions may propagate teacher errors to the student. Therefore, fine-grained distillation for autonomous driving should jointly answer two questions:
\emph{what knowledge is critical for the task}, and
\emph{whether the teacher is semantically consistent on that knowledge}.

Motivated by this observation, we propose \textbf{RSKD}, a
semantic-consistency-guided knowledge distillation framework for driving
VLMs. RSKD first localizes task-critical semantic phrases in the
ground-truth answer and evaluates the semantic consistency of the
corresponding teacher-predicted phrases. The resulting semantic consistency
score controls the balance between JSD distillation and ground-truth
supervision, emphasizing teacher signals that agree with the answer while
reducing the influence of inconsistent predictions.


Our main contributions are summarized as follows:
\begin{itemize}
    \item We introduce RSKD, a semantic-consistency-guided distillation framework for driving VLMs that explicitly identifies task-critical semantic phrases rather than treating all answer tokens uniformly.

    \item We propose a phrase-level semantic consistency mechanism that evaluates teacher predictions against the ground-truth semantics and adaptively routes JSD distillation and ground-truth supervision,
    strengthening teacher guidance only when it is semantically consistent.

    \item Experiments on DrivingVQA, LingoQA, and MapLMv2 show gains on the two QA benchmarks and competitive performance on MapLMv2 under the reported setting.
\end{itemize}

\section{Related Work}
\label{sec:related_work}

Knowledge distillation transfers knowledge from a large teacher to a compact student through distribution matching~\cite{hinton2015distilling}.
GKD~\cite{agarwal2024gkd} extends distribution-level distillation to autoregressive language models, while multimodal distillation methods such as LLAVADI~\cite{xu2024llavadi} investigate token- and logit-level knowledge transfer in multimodal models. These approaches mainly focus on improving teacher--student distribution or representation alignment, without explicitly identifying task-critical semantic phrases in driving answers.

Fine-grained methods further adapt supervision at the token level. ToDi~\cite{jung2025todi} and AdaKD~\cite{xie2026adakd} adjust knowledge transfer according to teacher--student divergence or token learning difficulty, while recent work also explores uncertainty-aware supervision ~\cite{sun2026betakd}. In contrast, RSKD localizes task-critical semantic phrases from the ground-truth answer, evaluates the semantic consistency of teacher predictions on these phrases, and uses the resulting signal to jointly route distribution distillation and ground-truth supervision.

\section{Method}
\label{sec:method}

\begin{figure*}[hptb]
    \centering
    \includegraphics[width=0.95\linewidth]{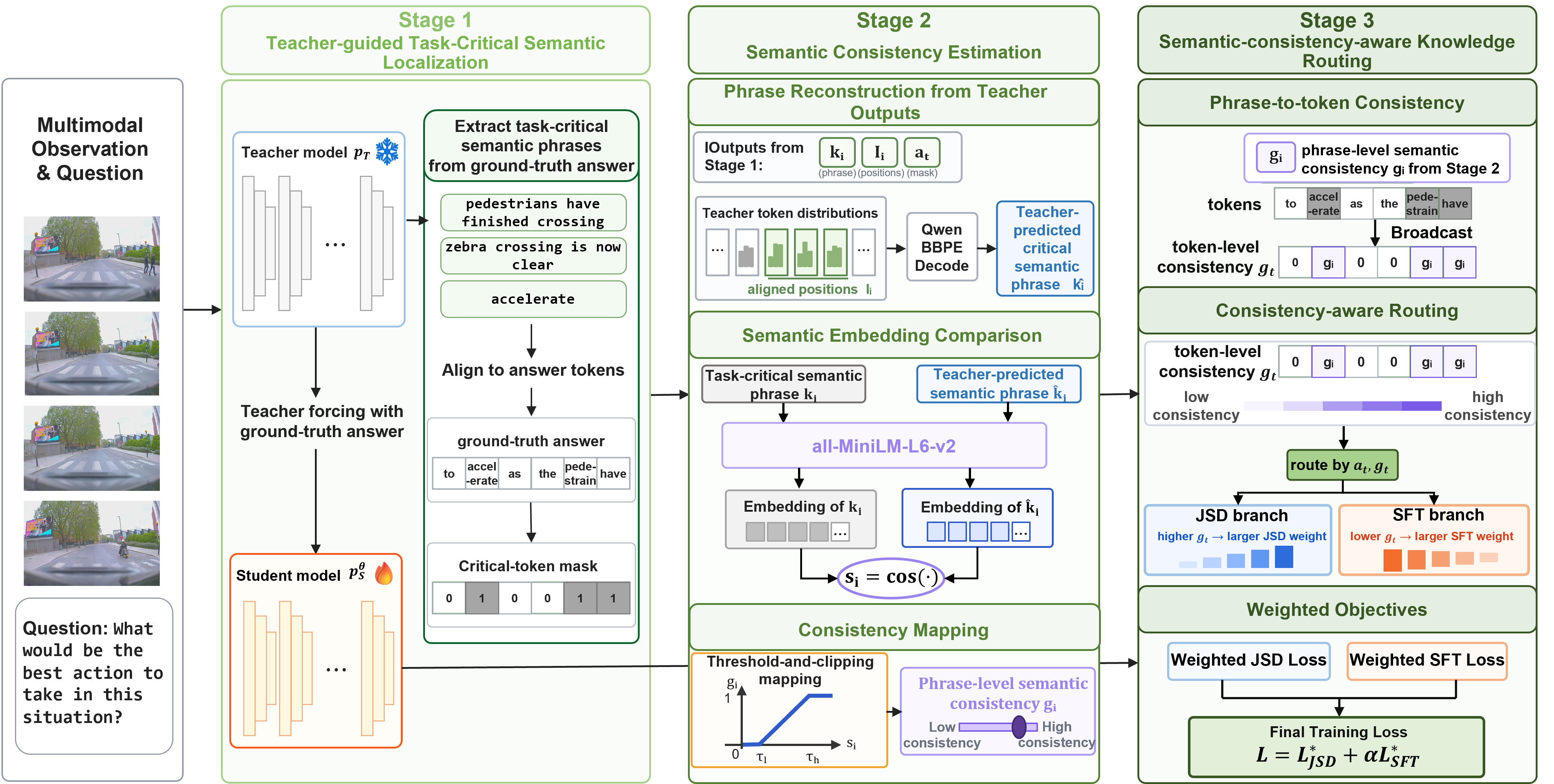}
    \caption{Overview of the proposed RSKD.}
    \label{fig:framework}
    \vspace{-13pt}
\end{figure*}

\subsection{Problem Formulation and Conventional Distillation}

Given an autonomous-driving multimodal question-answering instance, let
$x$ denote the multimodal input and $y=(y_1,\ldots,y_T)$ the ground-truth
answer. The teacher and student models are denoted by $p_T$ and
$p_S^\theta$, with next-token distributions

\begin{equation}
p_t(v)=p_T(v\mid x,y_{<t}), \qquad
q_t^\theta(v)=p_S^\theta(v\mid x,y_{<t}),
\quad v\in\mathcal{V},
\end{equation}

where $t\in\mathcal{T}$ indexes valid answer-token positions and
$\mathcal{V}$ is the vocabulary. Following GKD~\cite{agarwal2024gkd},
we use the generalized Jensen--Shannon divergence

\begin{equation}
\begin{aligned}
D_{\mathrm{JSD}}^\beta(p_t\|q_t^\theta)
&=
\beta D_{\mathrm{KL}}(p_t\|m_t)
+
(1-\beta)D_{\mathrm{KL}}(q_t^\theta\|m_t),\\
m_t
&=
\beta p_t+(1-\beta)q_t^\theta .
\end{aligned}
\end{equation}

Together with ground-truth supervision, the conventional objective is

\begin{equation}
\mathcal{L}_{\mathrm{joint}}
=
\frac{1}{|\mathcal{T}|}
\sum_{t\in\mathcal{T}}
\left[
D_{\mathrm{JSD}}^\beta(p_t\|q_t^\theta)
-
\alpha\log q_t^\theta(y_t)
\right],
\end{equation}

where $\alpha$ controls the ground-truth supervision strength.
Uniform token-level supervision, however, ignores both semantic importance
and teacher semantic consistency. RSKD addresses these factors through task-critical
semantic localization, consistency estimation, and consistency-aware routing.

\subsection{Framework Overview}

Figure~\ref{fig:framework} summarizes RSKD in three stages.
We first localize task-critical semantic phrases in the ground-truth answer, then estimate teacher semantic consistency on these phrases, and finally broadcast the phrase-level score to answer tokens to control JSD distillation and ground-truth supervision in opposite directions.

The main symbols are: $x$ for the multimodal observation and question, $y$
for the ground-truth answer, $\mathcal{K}$ for extracted task-critical semantic phrases,
$a_t$ for the critical-token mask, $g_i$ and $g_t$ for phrase- and token-level
semantic consistency scores, and $w_t^{\mathrm{JSD}}$ and
$w_t^{\mathrm{SFT}}$ for the routing weights. The teacher remains frozen throughout semantic extraction and distillation.

\subsection{Teacher-guided Task-Critical Semantic Localization}

The first step of our framework is to identify task-critical semantic phrases that are highly relevant to autonomous driving reasoning. Given the input $x$ and the ground-truth answer $y$, we prompt the teacher model to extract task-critical semantic phrases from the answer.




The extracted task-critical semantic phrase set is denoted as $\mathcal{K}=\{k_1,k_2,\ldots,k_N\}$, where each $k_i$ denotes a task-critical semantic phrase extracted from the ground-truth answer. Unlike answer generation, this process does not require the teacher model to generate a new answer. Instead, it performs semantic annotation over the known ground-truth answer, aiming to identify information that contributes most to driving-related reasoning.

To obtain token-level task-critical semantic localization, each phrase $k_i$ is aligned
with the tokenized answer sequence. Let
$\mathcal{I}_i=\mathcal{A}(k_i,y)$ denote the set of answer-token positions
aligned with $k_i$. The global critical-token mask is defined as

\begin{equation}
a_t=
\begin{cases}
1, & t\in\bigcup_{i=1}^{N}\mathcal{I}_i,\\
0, & \mathrm{otherwise},
\end{cases}
\end{equation}

where $\mathcal{A}(\cdot)$ aligns each phrase by exact token-subsequence
matching within the tokenized ground-truth answer.

The resulting mask $a_t$ provides a semantic prior indicating which answer tokens require consistency-aware supervision adjustment.

\subsection{Semantic Consistency Estimation}

For each task-critical semantic phrase $k_i$, we reconstruct the teacher prediction at its
aligned positions $\mathcal{I}_i$ under teacher forcing:
\begin{equation}
\hat{k}_i=
\mathrm{Dec}_{\mathrm{tok}}
\left(
\left[
\arg\max_{v\in\mathcal{V}} p_t(v)
\right]_{t\in\mathcal{I}_i}
\right),
\end{equation}
where $\mathrm{Dec}_{\mathrm{tok}}$ denotes decoding with the Qwen3 tokenizer BBPE~\cite{bai2025qwen3vl}.

We then encode $k_i$ and $\hat{k}_i$ using a frozen all-MiniLM-L6-v2
sentence encoder~\cite{reimers2019sentencebert,wang2020minilm}. Their semantic
similarity and consistency score are
\begin{equation}
s_i=
\cos\!\left(\phi(k_i),\phi(\hat{k}_i)\right),\qquad
g_i=
\mathrm{clip}\!\left(
\frac{s_i-\tau_l}{\tau_h-\tau_l},0,1
\right).
\end{equation}


A larger $g_i$ indicates stronger semantic agreement between the teacher prediction and the ground-truth phrase, making its predictive distribution more reliable for supervising student distillation; a smaller $g_i$, in contrast, calls for greater reliance on ground-truth supervision. Explicit contradictions between mutually exclusive driving relations are assigned $g_i=0$. The phrase-level score is then broadcast to the aligned token positions as $g_t=g_i$ for $t\in\mathcal{I}_i$.

\subsection{Semantic-consistency-aware Knowledge Routing}

Given the critical-token mask $a_t$ and consistency score $g_t$, we route
JSD and ground-truth supervision in opposite directions:
\begin{equation}
\begin{aligned}
w_t^{\mathrm{JSD}}
&=1-a_t+a_t
\left[
w_{\min}^{\mathrm{JSD}}
+
\left(w_{\max}^{\mathrm{JSD}}-w_{\min}^{\mathrm{JSD}}\right)g_t
\right],\\
w_t^{\mathrm{SFT}}
&=1-a_t+a_t
\left[
w_{\min}^{\mathrm{SFT}}
+
\left(w_{\max}^{\mathrm{SFT}}-w_{\min}^{\mathrm{SFT}}\right)(1-g_t)
\right].
\end{aligned}
\end{equation}
Thus, critical tokens with high semantic consistency receive stronger JSD
supervision, while those with low consistency receive stronger ground-truth
supervision.

The weighted objectives and final loss are
\begin{equation}
\begin{aligned}
\mathcal{L}_{\mathrm{JSD}}^{*}
&=
\frac{
\sum_{t\in\mathcal{T}}
w_t^{\mathrm{JSD}}
D_{\mathrm{JSD}}^\beta(p_t\|q_t^\theta)
}{
\sum_{t\in\mathcal{T}} w_t^{\mathrm{JSD}}
},\\
\mathcal{L}_{\mathrm{SFT}}^{*}
&=
-\frac{
\sum_{t\in\mathcal{T}}
w_t^{\mathrm{SFT}}\log q_t^\theta(y_t)
}{
\sum_{t\in\mathcal{T}} w_t^{\mathrm{SFT}}
},\\
\mathcal{L}
&=
\mathcal{L}_{\mathrm{JSD}}^{*}
+
\alpha\mathcal{L}_{\mathrm{SFT}}^{*}.
\end{aligned}
\end{equation}

Each branch is normalized by its own weight sum, so routing changes the
relative emphasis within each supervision source rather than enforcing a fixed
global JSD-to-SFT ratio.




\section{Experiments}

\subsection{Experimental Setup}

\textbf{Benchmarks and baselines.}
We evaluate on DrivingVQA~\cite{corbiere2026drivingvqa},
LingoQA~\cite{marcu2024lingoqa}, and MapLMv2~\cite{cao2024maplm}
using their official metrics. We report the frozen teacher and unadapted
student as references, and compare against SFT, off-policy
GKD~\cite{agarwal2024gkd}, ToDi~\cite{jung2025todi}, and RSKD.
All trained student methods use the same training mixture of DrivingVQA,
LingoQA, and MapLMv2.

\textbf{Implementation details.}
We use Qwen3-VL-4B-Instruct as the student and an internally fine-tuned XEmbodied teacher~\cite{qian2026xembodied} based on Qwen3-VL-30B-A3B-Instruct~\cite{bai2025qwen3vl}.
The student is adapted with LoRA~\cite{hu2022lora} on all linear language-model layers with rank 8, scaling factor 32, and dropout 0.05, while the visual encoder and aligner are frozen. We train for two epochs with BF16, a learning rate of $1\times10^{-5}$, per-device batch size 1, no gradient accumulation, warmup ratio 0.05, and maximum sequence length 32,000. Each trained configuration is repeated with $3$ random seeds, and we report the mean performance across runs. Training is distributed over 96 NVIDIA H20 GPUs, averaging 52h 20min per run.

For task-critical semantic phrase extraction, the teacher is prompted to return exact contiguous substrings from the ground-truth answer describing traffic participants, states, spatial relations, road constraints, or required behaviors. These phrases are aligned to answer tokens by exact token-subsequence matching; empty, malformed, or unmatched phrases are ignored. Teacher predictions that contradict the ground-truth phrase on predefined mutually exclusive driving relations (e.g., left/right, front/behind, moving/stopped, or accelerate/decelerate) are assigned zero consistency.


We set $\tau_l=0.6$ and $\tau_h=0.8$ using the validation split and use JSD/SFT weight ranges of $[0,2]$/$[2,4]$. We further select the loss coefficient $\alpha$, JSD mixing coefficient $\beta$, and distillation
temperature on the validation split by searching over $\alpha\in\{0.1,0.5,1,2\}$, $\beta\in\{0.1,0.5,0.9\}$, and temperatures $\{1,2,4\}$, yielding $\alpha=1$, $\beta=0.5$, and temperature $1$. No test data are used for hyperparameter selection.




\begin{table}[t]
\centering
\caption{Main results on three autonomous-driving vision-language benchmarks. Higher is better. Trained student results are averaged over multiple random seeds; the best student-based result is bold.}
\label{tab:main_results}

\setlength{\tabcolsep}{2pt}
\renewcommand{\arraystretch}{1.05}

\begin{tabularx}{\columnwidth}{
@{}
>{\raggedright\arraybackslash}p{0.22\columnwidth}
>{\centering\arraybackslash}X
>{\centering\arraybackslash}X
>{\centering\arraybackslash}X
@{}
}
\hline
\textbf{Method}
& \textbf{DrivingVQA~$\uparrow$}
& \textbf{LingoQA~$\uparrow$}
& \textbf{MapLMv2~$\uparrow$} \\
\hline

Teacher
& 49.81
& 0.642
& 79.00 \\

Student
& 3.30
& 0.584
& 53.26 \\

\hline

SFT
& 14.06
& 0.631
& \textbf{81.85} \\

GKD
& 16.34
& 0.640
& 81.08 \\

ToDi~\cite{jung2025todi}
& 16.35
& 0.649
& 81.80 \\

RSKD
& \textbf{16.60}
& \textbf{0.672}
& 81.82 \\

\hline
\end{tabularx}
\vspace{-13pt}
\end{table}












\subsection{Main Results}




Table~\ref{tab:main_results} reports the main results. RSKD achieves the best student performance on DrivingVQA and LingoQA, reaching 16.60 and 0.672, respectively, and consistently outperforming SFT, GKD, and ToDi. The improvements over GKD and ToDi suggest that explicitly incorporating task-critical semantic localization and teacher--ground-truth consistency can complement uniform or divergence-based distillation. On MapLMv2, the teacher itself performs relatively weakly (79.00), limiting the useful knowledge available for transfer. RSKD nevertheless reaches 81.82, substantially above the teacher and GKD and nearly matching SFT (81.85), which is consistent with its design of reducing reliance on semantically inconsistent teacher signals while strengthening ground-truth supervision.

\subsection{Ablation Study}

For efficiency, all ablations use the same 50\% LingoQA subset under an identical reduced training budget, and Table~\ref{tab:ablation} is compared only within this setting. Random Spans replaces task-critical spans with random answer spans, while w/o Consistency removes consistency estimation using fixed semantic weights, and w/o SFT Routing disables consistency-aware routing in the ground-truth supervision branch. Replacing task-critical spans with random ones reduces the score from 0.480 to 0.474, confirming the value of semantic localization, while removing consistency estimation yields 0.475, showing that semantic importance alone is insufficient. Removing SFT routing causes the largest drop to 0.468, highlighting the importance of stronger ground-truth supervision when teacher predictions are semantically inconsistent.
\begin{table}[t]
\centering
\caption{Ablation on a fixed 50\% LingoQA subset under the same reduced training budget.}
\label{tab:ablation}
\setlength{\tabcolsep}{2.4pt}
\renewcommand{\arraystretch}{1.05}
\begin{tabular}{@{}lccccc@{}}
\hline
\textbf{Variant} &
\textbf{Loc.} &
\textbf{Cons.} &
\textbf{JSD Rt.} &
\textbf{SFT Rt.} &
\textbf{Score $\uparrow$} \\
\hline
Random Spans
& Rand. & \checkmark & \checkmark & \checkmark & 0.474 \\
w/o Consistency
& \checkmark & -- & -- & -- & 0.475 \\
w/o SFT Routing
& \checkmark & \checkmark & \checkmark & -- & 0.468 \\
Full RSKD
& \checkmark & \checkmark & \checkmark & \checkmark & \textbf{0.480} \\
\hline
\end{tabular}
\vspace{-13pt}
\end{table}
\subsection{Limitations}

RSKD has several limitations. Its semantic consistency estimation depends on teacher phrase extraction and a frozen MiniLM encoder, making the routing sensitive to extraction and embedding errors. The fixed similarity thresholds and rule-based contradiction detection may also require recalibration across datasets with different response styles. In addition, RSKD introduces extra computation during training, while our evaluation is currently limited to a single teacher--student pair; broader validation is left for future work.

\section{Conclusion}

We present RSKD, a semantic-consistency-guided distillation framework for driving VLMs that localizes task-critical semantic phrases and routes teacher and ground-truth supervision according to semantic similarity. Experiments on DrivingVQA, LingoQA, and MapLMv2 show gains over distillation baselines on the two QA benchmarks and competitive performance when the teacher advantage is limited, highlighting the need to identify task-critical semantics and assess whether teacher predictions preserve their ground-truth meaning.

\vfill\pagebreak


\bibliographystyle{IEEEbib}
\bibliography{strings,refs}

\end{document}